\documentclass{llncs}

\usepackage[utf8]{inputenc} 
\usepackage[T1]{fontenc}    
\usepackage{hyperref}       
\usepackage{url,multirow}            
\usepackage{booktabs}       
\usepackage{amsfonts}       
\usepackage{nicefrac}       
\usepackage{microtype}      
\usepackage{lipsum,graphicx,amsmath}
\begin{document}

\title{Human-to-AI Coach:\\ Improving Human Inputs to AI Systems}

\author{Johannes Schneider\inst{1}}
\institute{Institute of Information Systems, University of Liechtenstein, Liechtenstein  }

\maketitle

\begin{abstract} 
Humans increasingly interact with Artificial intelligence(AI) systems. AI systems are optimized for objectives such as minimum computation or minimum error rate in recognizing and interpreting inputs from humans. In contrast, inputs created by humans are often treated as a given. We investigate how inputs of humans can be altered to reduce misinterpretation by the AI system and to improve efficiency of input generation for the human while altered inputs should remain as similar as possible to the original inputs. These objectives result in trade-offs that are analyzed for a deep learning system classifying handwritten digits. To create examples that serve as demonstrations for humans to improve, we develop a model based on a conditional convolutional autoencoder (CCAE). Our quantitative and qualitative evaluation shows that in many occasions the generated proposals lead to lower error rates, require less effort to create and differ only modestly from the original samples.
\end{abstract}

\section{Introduction} 
\begin{figure*}[!ht]
\vspace{-6pt}
	\centering
	\includegraphics[width=1\textwidth]{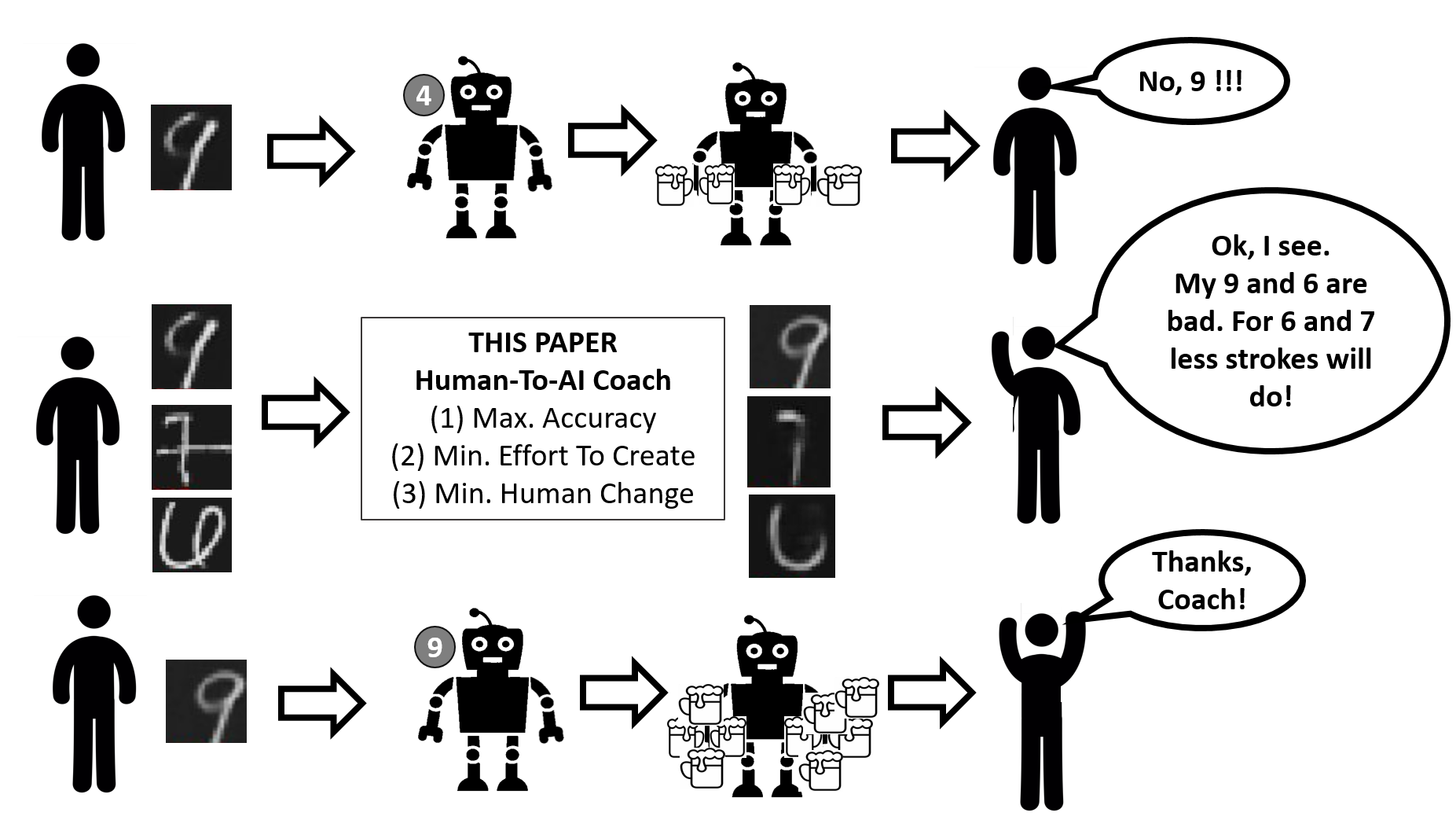}
	\caption{``Human-to-AI'' (H2AI) coach: From misunderstandings to understanding} \label{fig:comm}
	\vspace{-12pt}
\end{figure*}

Human-to-AI information flow is increasing rapidly in importance and extent across multiple modalities. For example, voice-machine interaction is becoming more and more popular with deep learning networks recognizing text from speech. Similar, the progress in image recognition has lowered error rates in gesture and optical character recognition. Still, key technologies in AI such as deep learning are not perfect. They might also err given ambiguous inputs created by humans. Errors might be more likely by humans being in a hurry, being unaware of the AI's recognition mechanism, sloppiness or lack of skill. Safety critical application areas such as autonomous driving or medical applications, where an AI might depend on inputs from humans in one way or another, are becoming more and more prominent. Thus, mistakes in recognizing and processing inputs should be avoided. Apart from avoiding errors, humans might also have an incentive to provide inputs with less effort, eg. ``Why try to speak clearly and loudly in the presence of noise, if mumbling works just as well? Why doing that extra stroke in writing a character, if detection works just as well without it?''  In this work, we do not focus on how to improve AI systems that recognize and interpret human information. We aim at strategies how humans can convey information better to such a system by adjusting their behavior. Identifying potential improvements becomes more difficult when deep learning is involved. Improvements are often based on a deep understanding of mechanisms of the task at hand, ie. how an AI system processes inputs. Deep learning is said to follow a black-box behavior. Even worse, deep learning is well-known to reason very differently from humans: Deep learning models might astonish due to their high accuracy rates, but disappoint at the same time by failing on simple examples that were just slightly modified as well-documented by so called ``adversarial examples''. As such, humans might depend even more on being shown opportunities for generating better data that serves as input to an AI. In this work, we formalize the aforementioned  partially conflicting goals such as minimizing wrongly recognized human inputs and reducing effort for humans -- both in terms of need to adjust their behavior as well as to interact effortlessly. We focus on the classification problem of digits, where we aim to provide suggestions to humans by altering their generated inputs as illustrated in Figure \ref{fig:comm}. We express the problem in terms of a multi-objective optimization problem, ie. as a linear weighted sum. As model we use a conditional convolutional autoencoder. Our qualitative and quantitative evaluation highlights that the generated samples are visually appealing, easy to interpret and also lead to a lower error rate in recognition.

\section{Challenges of Human-to-AI Communication}
We consider the problem of improving human generated inputs to an AI illustrated in Figure \ref{fig:comm}. A human wants to convey information to an AI using some mode, eg. speech, writing, or gestures. The processing of the received signals by the AI often involves two steps: (i) recognition, ie. identifying and extracting relevant information in the input signal, and (ii) interpretation, ie. deriving actions by utilizing the information in a specific context. For recognition, the information has to be extracted from a physical (analog) signal, eg. using speech recognition, image recognition, etc. In case information is communicated in a digital manner using structured data, recognition is commonly obsolete. Often the extracted information has to be further processed by the AI using some form of sense-making or interpretation. The AI requires potentially semantic understanding capabilities and might rely on the use of context such as prior discourse or surrounding. We assume that the human interacts frequently with such a system, so that it is reasonable for the human to improve on objectives such as errors and efficiency in communication. In this paper, we consider the challenge of discovering variations of the original inputs that might help a human to improve. \\
More formally, we consider a classification problem, where a user provides data $D=(X,Y)$. Each sample $X$ should be recognized as class $Y$ by a classifier $C_H$. We denote by $X_i$ the $i$-th feature of sample $X$. For illustration, for the case of handwritten digits a sample $X$ is a gray-tone scan of a digit and $Y \in [0-9]$ the digitized number. $X_i \in [0,1]$ gives the brightness of the $i$-th pixel in the scan. The classification model $C_H$ was trained to optimize classification performance of human samples, ie. maximize $P_{C_H}(Y|X)$. We regard the model $C_H$ as a given, ie. we do not alter it in any way, but use it in our optimization process. The Human-to-AI coach ``H2AI'' takes as input one sample $X$ with its label $Y$. It returns at least one proposal $\hat{X}$, ie. $\hat{X}:=H2AI(X,Y)$. The suggestion $\hat{X}$ should be superior to $X$ according to some objective, eg. we might demand higher certainty in recognition $P_{C_H}(Y|X)<P_{C_H}(Y|\hat{X})$. In a handwriting scenario a human might use a proposal $\hat{X}$ based on an input $X$ to adjust her strokes. 

\section{Model and Objectives} 

An essential requirement is that the modified samples are similar to the given input, otherwise a trivial solution is to always return ``the perfect sample'' that is the same for any input. This motivates utilizing an auto-encoder (Section \ref{sec:arch}) and adding multiple loss terms to handle various objectives (Section \ref{sec:loss}).

\subsection{Architecture} \label{sec:arch}
\begin{figure*}[!ht]
\vspace{-6pt}
\setlength\abovecaptionskip{-0.05\baselineskip}
\setlength\belowcaptionskip{-1.0\baselineskip}
	\centering
	\includegraphics[width=0.9\textwidth]{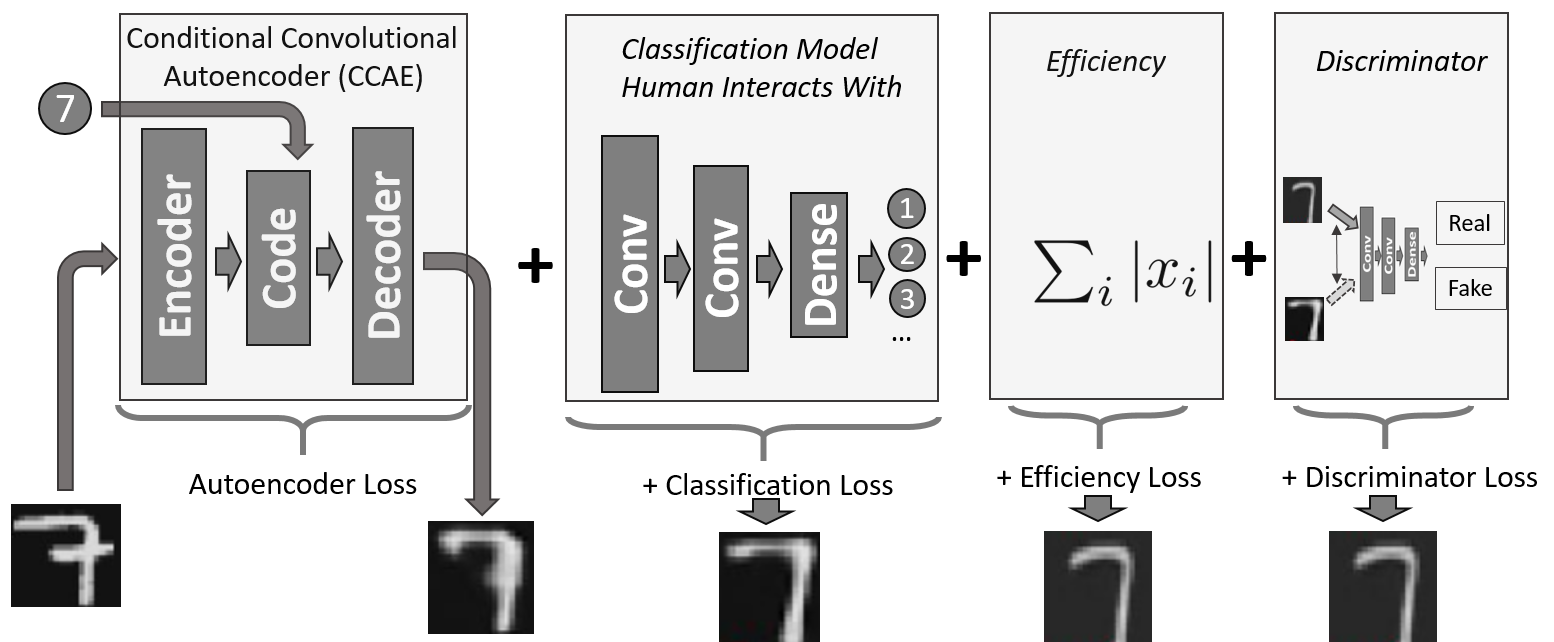}
	\caption{Human-to-AI(H2AI) model with its components and regularizers} \label{fig:mod}
	\vspace{-24pt}
\end{figure*}

Two approaches that allow to create (modified) samples are generative adverserial networks(GANs) and autoencoders(AEs). There are also combinations thereof, eg. the pix2pix architecture \cite{iso17} or conditional variational autoencoder \cite{bao17}. \cite{iso17} and \cite{bao17} contain an AE which has a decoder serving as a generator based on a latent representation from the encoder and, additionally, a discriminator. AE tend to generate outcomes that are closer to the inputs. But they are often smoother and less realistic looking. In our application staying close to the input is a key requirement, since we only want to show how a sample can be modified rather than generating completely new samples. Thus, we decided to focus on an AE-based architecture. We also investigate including a discriminator to improve generated samples. More precisely, we utilize conditional AE with extra loss terms for regularization covering not only a discriminator loss but also losses for efficiency and classification of modified samples as shown in Figure \ref{fig:mod}. Conditional AE are given as input the class of a sample in addition to the sample itself. This often improves generated samples, in particular for samples that are ambiguous, ie. samples that seem to match multiple classes well.  

\begin{figure*}[!ht]
    \setlength\abovecaptionskip{-0.05\baselineskip}
    \setlength\belowcaptionskip{-1.0\baselineskip}
	\centering
	\includegraphics[width=0.75\textwidth]{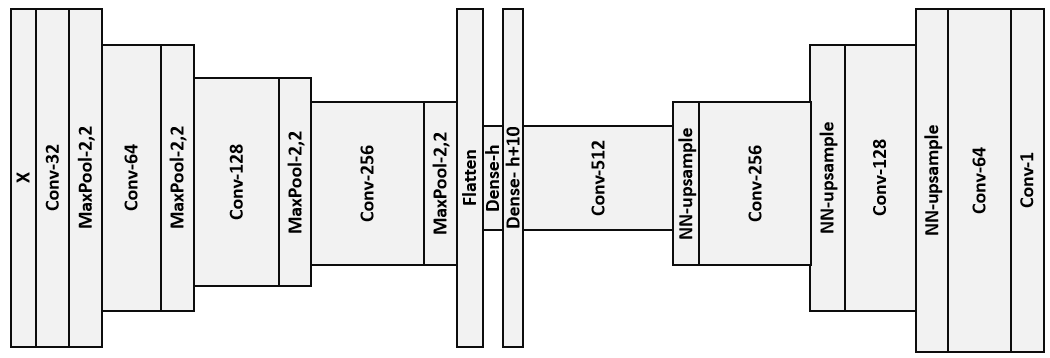}
	\caption{H2AI implementation using a convolutional conditional autoencoder (CCAE)} \label{fig:ccae}
	\vspace{-6pt}
\end{figure*}

Convolutional AE are known to work well on image data. Therefore, we propose convolutional conditional AE (CCAE) as shown in Figure \ref{fig:ccae}, where the NN-upsample layers in the decoder denote nearest-neighbor upsampling. After each convolutional layer, there is a ReLU layer that is not shown in Figure \ref{fig:ccae}. Compared to transposed convolutional layers, NN-upsampling with convolutional layers prevents checkerboard artifacts in the resulting images.

\subsection{Objectives and Loss Terms}\label{sec:loss}
The generated input samples should meet multiple criteria, each of which is implemented as a loss term. The loss terms and their weighted sum (with parameters $\alpha_{\cdot}$) are given in Equations \ref{eq:los} and illustrated in Figure \ref{fig:mod}. The total loss $L_{Tot}(X,Y)$ contains four parameters $\alpha_{RE}$, $\alpha_{CL}$, $\alpha_{EF}$ and $\alpha_{D}$. It is possible to keep $\alpha_{RE}$ and use the other three to control the relative importance of the following objectives:
\begin{equation} 
	\footnotesize
	\vspace{-3pt}
	\begin{aligned}	
	&\hat{X}:=CCAE(X,Y) \text{\phantom{abc}Sample proposed by H2AI-coach} \\
	 &L_{RE}(X,\hat{X}) := \sum_i |X_i-\hat{X}_i| \text{\phantom{abc}Reconstruction or Change Loss} \label{eq:los}\\
	 &L_{CL}(\hat{X},Y) \text{\phantom{abc}Classification Loss} \\ 
	 &L_{EF}(\hat{X}) := {\sum_i |\hat{X}_i|} \text{\phantom{abc}Efficiency Loss}\\
	 &L_{D}(\hat{X}) := \log(1 - D(\hat{X})) \text{\phantom{abc}Discriminator Loss} \\ 
	 &L_{Tot}(X,Y) :=  \alpha_{RE} L_{RE}(X,\hat{X})+\alpha_{CL}L_{CL}(\hat{X},Y)  +\alpha_{EF}  L_{EF}(\hat{X}) +\alpha_{D} L_{D}(\hat{X})\\
	\end{aligned}	
\end{equation} 
\noindent \textbf{Minimal effort to change}: Change might be difficult and tedious for humans. Thus, the effort for humans to adjust their behavior should be minimized. This implies that the original samples $X$ created by humans and the newly generated variations $\hat{X}$ should be similar. This is covered by the reconstruction loss $L_{RE}(X,\hat{X})$ of the AE (see Equations (\ref{eq:los})). It enforces the output and the input to be similar. But parts of the input might be changed fairly drastically, ie. for handwritten digits pixels might change from 0(black) to 1(white) and vice versa. For that reason, we do not employ an $L2$-metric, which heavily penalizes such differences, but rather opt for an $L1$-metric.\\
\noindent \textbf{Reduce mis-understanding}: The amount of wrongly extracted or interpreted information by the AI should be reduced. AEs are known to have a denoising, averaging effect. They are also known to improve performance in some cases in conjunction with classification tasks \cite{mak13}. To further foster a reduction in mis-understandings we minimize the classification loss $L_{C_H}(\hat{X},Y)$ for generated examples $\hat{X}$ for the model $C_H$ the human communicates with.\\
\noindent \textbf{Realistic samples}: The generated samples $\hat{X}$ should still be comprehensible for humans or other systems, ie. look realistic. It can happen that a generated proposal $\hat{X}$ is so optimized for the given AI model $C_H$ that it is not meaningful in general. That is, the proposal $\hat{X}$ might appear not only very different from prototypical examples of its class but very different from any example occurring in reality. While AEs partially counteract this, AEs do not enforce that samples look real, but tend to create smooth (averaged) samples. Thus, we add a discriminator $D$ resulting in a GAN architecture that should distinguish between real and generated samples and make them look crispier. The added discriminator loss $L_{D}(\hat{X})$ is $\log(1 - D(\hat{X}))$, where $\hat{X}$ is the generated sample $\hat{X}:=CCAE(X,Y)$ for an input sample $X$ of a human of class $Y$.\\
\noindent \textbf{Minimal effort to create samples}: Interaction should be effortless for the human (and AI). To quantify effort of a human to create a sample, time might be a good option if available. If not, application specific measures might be more appropriate. For measuring effort in handwriting, the amount (and length) of strokes can be used. A good approximation can be the total amount of needed ``ink'', which corresponds to the $L1$-loss of the proposal $\hat{X}$, ie.  $L_{EF}(\hat{X}) := {\sum_i |\hat{X}_i|}$. We chose the $L1$ over the $L2$-metric, since having many low intensity pixels (as fostered by $L2$) is generally discouraged.

\section{Evaluation}
We conducted both a qualitative and quantitative evaluation on the MNIST dataset, since it has been used by recent work in similar contexts \cite{goy19,dhu18}. It consists of 50000 handwritten digits from 0 to 9 for training and 10000 digits for testing. 
The classification model $C_H$, ie. the system a user is supposed to communicate well with, is a simple convolutional neural network (CNN) consisting of two convolutional layers (8 and 16 channels) that are both followed by a ReLU and 2x2 Max-Pooling Layer. The last layer is a fully connected layer. The network achieved a test accuracy of 95.97$\%$. While this could be improved, it is not of prime relevance for our problem, since the classifier $C_H$ is treated as a given. The architecture of the H2AI coach is shown in Figure \ref{fig:mod} with details of the AE in Figure \ref{fig:ccae} and loss terms in Equations \ref{eq:los}.  We did not employ any data augmentation. We used the AdamOptimizer with learning rate 1e-4 for all models. Training lasted for 10 epochs with a batchsize of 8. We trained 5 networks for each hyperparameter setting. We perform statistical analysis of our results using t-tests.

For the ablation study we consider adding each of the losses in isolation to the baseline with just the AE by varying parameters $\alpha_{CL},\alpha_{EF},\alpha_{D}$ that control their impact. For the AE we used $\alpha_{RE}=32$ for all experiments.\footnote{$\alpha_{RE}$ is not needed (could be set to 1). But, in practice, it is easier to vary $\alpha_{RE}$ than changing $\alpha_{CL}, \alpha_{EF},\alpha_{D}$ since they behave non-linearly.} Finally, we consider a model, where we add all losses. There are no fixed ranges for the parameters $\alpha$, but they should be chosen so that all loss terms have a noticeable impact on the total loss -- at least in the early phases of training.\footnote{We found that altering $\alpha$ during training requires much more tuning, but yields only modest improvements.}

Our qualitative analysis is a visual assessment of the generated images. We investigate images that were improved (in terms of each of the metrics), worsened and remained roughly the same. As quantitative measures we used the losses as defined in Equations \ref{eq:los} except for classification, where we used the more common accuracy metric. 

\subsection{Qualitative Analysis}
\begin{figure}[!ht]
\vspace{-6pt}
    \setlength\abovecaptionskip{-0.05\baselineskip}
    \setlength\belowcaptionskip{-1.0\baselineskip}
	\centering \includegraphics[width=0.85\textwidth]{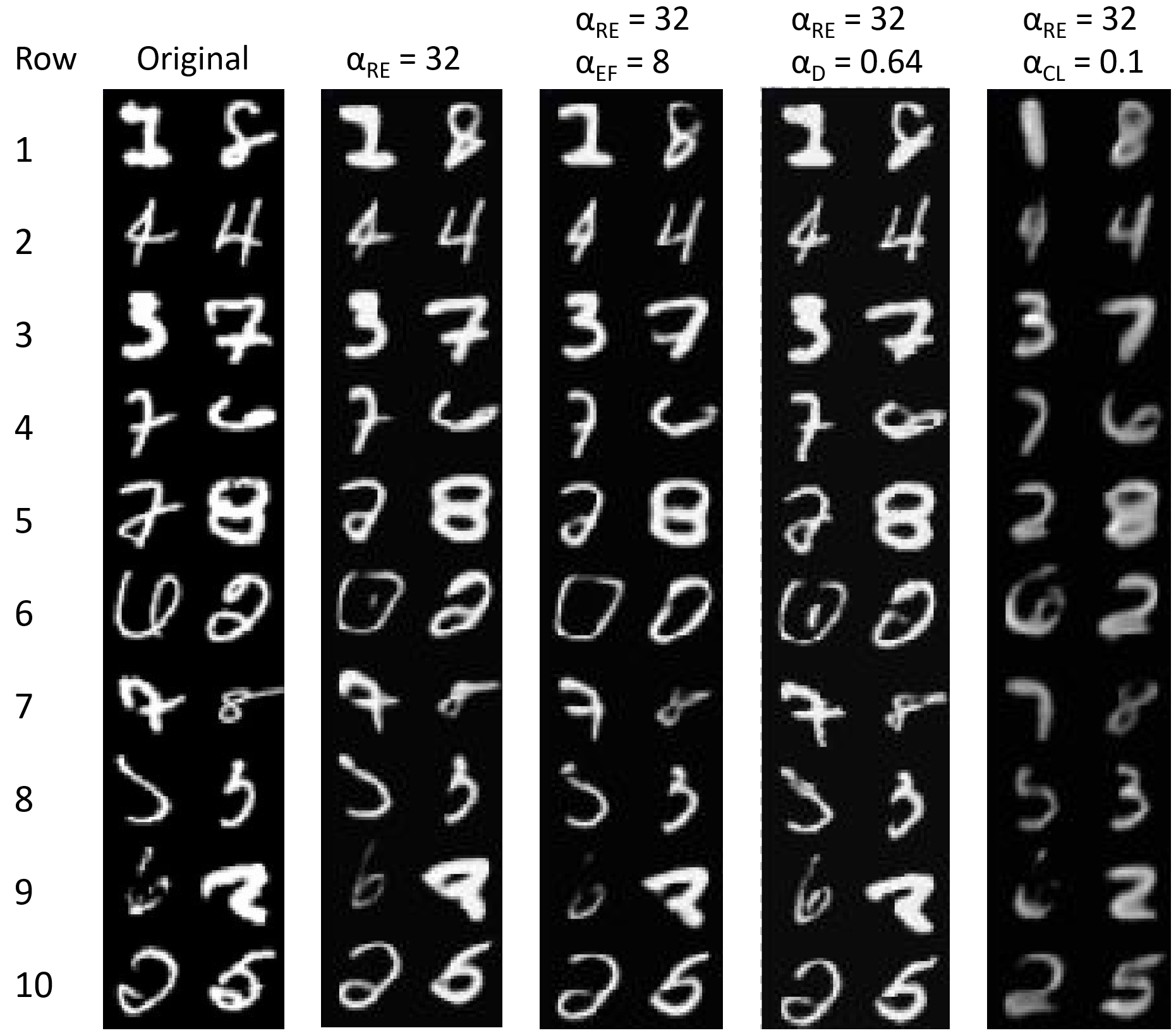}
	\caption{Original and generated samples using a subset of all loss terms} \label{fig:clm}
\end{figure}

Figure \ref{fig:clm} shows unmodified samples (left most column) and various configurations of loss weights $\alpha$. We use \textit{R.x} to denote ``row x''.
The AE (2nd column, $\alpha_{RE}=32$) on its own already has overall a positive impact yielding smoother images than the original ones. It tends to improve efficiency by removing ``exotic'' strokes, eg. for the \textit{2} in R.6 and the \textit{5} in the last row, and sometimes helps also in improving readability (eg. ease of classification), eg. the \textit{8} in R.1 row and the \textit{6} in the 2nd last row both become more readable. Other digits might seem more readable but are actually worsened, eg. the \textit{6} in R.6 appears to become a \textit{0} (it is actually a \textit{6}) and the \textit{7} in R.7 appears to become more of a \textit{9}.  When optimizing in addition for efficiency (3rd column), some parts of digits get deleted, which is sometimes positive and sometimes negative. Some benefits of the AE seem to get undone, eg. the \textit{6} in the 2nd last row now looks again more like the original with missing parts. The same holds for the \textit{8} in R.1, though for both some improvement in shape remains. More interestingly, the digits in R.6 both get changed to \textit{0}, which is incorrect. On the positive side, several figures become more readable through subtle changes, eg. removals of parts like the \textit{5} in the last row, the \textit{2} in the 2nd last row or the \textit{3} in R.3. When using the AE and the discriminator (4th column in Figure \ref{fig:clm}), we can observe that the samples become slightly more realistic, ie. crispier. We can see clear improvements for the \textit{7} in R.7 and the \textit{6} in R.9. Many digits remain the same. When using the AE and the classification loss (last column) smoothness increases and digits appear blurry. Readability worsens for a few digits, ie. the left \textit{4} in R.2 can now be easily confused with a \textit{9}, the \textit{6} in R.9 is no better than the original and worse than the one using a discriminator. Overall, the classification loss helps to improve many other samples. Some only now become readable, eg. the \textit{5} and \textit{3} in R.8. Also some digits become simpler, eg. the \textit{1} R.1 and the \textit{7}s in R.3, R.4 and R.7.

\begin{figure}[!ht]
\vspace{-12pt}
    \setlength\abovecaptionskip{-0.05\baselineskip}
    \setlength\belowcaptionskip{-1.0\baselineskip}
	\centering \includegraphics[width=0.85\textwidth]{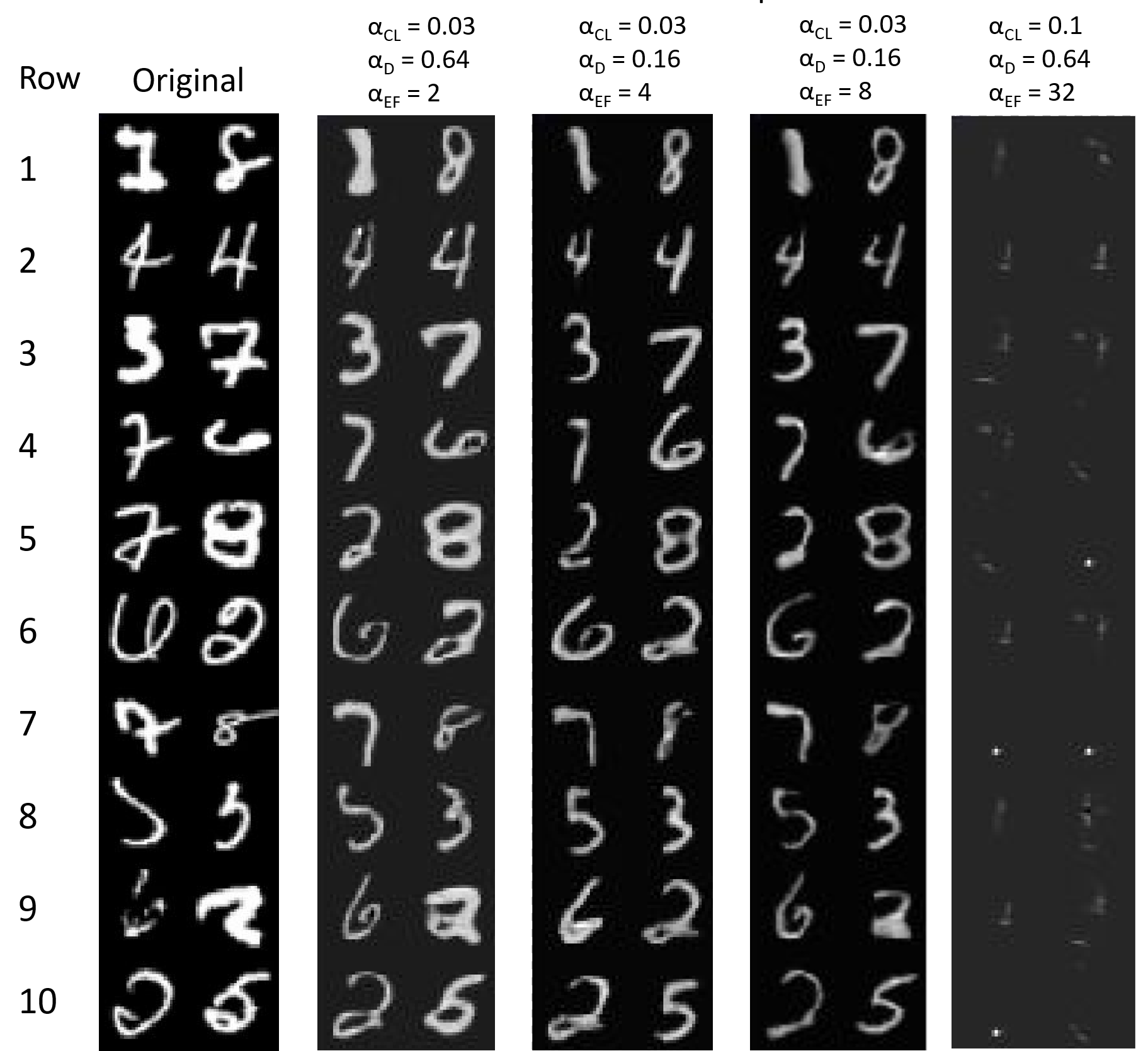}
	\caption{Original and generated samples using all loss terms} \label{fig:clmall}
	\vspace{-12pt}
\end{figure}

When combining all losses (Figure \ref{fig:clmall}) it can be observed that for some parameters $\alpha$ larger values are possible to get reasonable results, since the objectives might counteract each other. For example, the discriminator loss pushes pixels to become brighter, whereas the efficiency loss pushes them to be darker. We noticed that the strong smoothing effect due to the classification loss is essentially removed mainly due to the discriminator loss but also partially due to the efficiency loss. The benefits of the classification loss, however, mainly remain and are also improved: The \textit{4} in the R.2 and the \textit{6} in R.9 become more readable. There are also differences in quality among the three configurations. Interestingly, the original images show somewhat more contrast, in particular compared to the second column. A careful observer will notice a few bright points in the upper part of both \textit{4} in R.2. These seem to be artifacts of the optimization. It is well-known that training GANs might lead to non-convergence or mode-collapse. The former was observed for (too) large discriminator loss $\alpha_{D}$. We also noticed mode collapse for large values of $\alpha_{CL}$ (not shown) and bad outcomes for large values of $\alpha_{EF}$ as shown in the last column. Degenerated examples still score high in some of the metrics, but are very poor in others, eg. in the last column accuracy and efficiency loss are good, but reconstruction loss is large.  Still, overall combining all losses leads to best results.

\subsection{Quantitative Analysis}
Table \ref{tab:res} shows the loss terms (with accuracy instead of classification loss) for all loss configurations also shown in Figure \ref{fig:clm} for our ablation study with the reconstruction loss (AE only) as baseline.
\begin{table*}[!htb]
    \setlength\tabcolsep{4.5pt}
	\centering
	\begin{tabular}{|c|c|c| c|c|c| c|		} \hline
	Loss &$\alpha_{CL}$ &$\alpha_{EF}$ &$\alpha_{D}$ & $Accuracy$& $L_{RE}$ & $L_{EF}$\\ \hline 
Baseline(AE only) &0.0&0.0&0.0&0.9609&0.00018&0.00097\\ \hline
\multirow{4}{*}{\centering Classific. Loss}&0.03&0.0&0.0&0.9994&0.00027&0.00096\\ \cline{2-7}
&0.08&0.0&0.0&0.9997&0.00041&0.00096\\ \cline{2-7}
&0.1&0.0&0.0&0.9998&0.00042&0.00092\\ \cline{2-7}
&0.24&0.0&0.0&1.0&0.00062&0.00085\\ \hline
\multirow{5}{*}{\centering Efficiency Loss}&0.0&1.0&0.0&0.9587&0.00019&0.00095\\ \cline{2-7}
&0.0&4.0&0.0&0.9607&0.00018&0.00093\\ \cline{2-7}
&0.0&8.0&0.0&0.9578&0.00019&0.00091\\ \cline{2-7}
&0.0&16.0&0.0&0.9458&0.00023&0.00081\\ \cline{2-7}
&0.0&32.0&0.0&0.1135&0.00098& $<$1e-5\\ \hline
\multirow{3}{*}{\centering Discrim. Loss}&0.0&0.0&0.03&0.9608&0.00019&0.00099\\ \cline{2-7}
&0.0&0.0&0.16&0.96&0.0002&0.00096\\ \cline{2-7}
&0.0&0.0&0.64&0.9318&0.00032&0.00096\\ \hline
	\end{tabular}		
	\caption{Results varying one loss term weight $\alpha_{CL}$, $\alpha_{EF}$,$\alpha_{D}$}	\label{tab:res}	
	\vspace{-3pt}
\end{table*}
We first discuss accuracy. The AE on its own leads to a small gain in accuracy compared to the baseline classifier $C_H$ of 95.97$\%$.  Not surprisingly, optimizing accuracy directly (using a classification loss, ie. $\alpha_{CL}>0$) leads to best results: even for a seemingly small $\alpha_{CL}$ accuracy exceeds .999$\%$. While it appears that differences in accuracy between various values of $\alpha_{CL}$ are not significant, from a statistical perspective (using a t-test) they are (p-value $<$ .001). For any $\alpha_{CL}$, the network tends to always fail to learn the same samples, leading to very low variance in accuracy. The large accuracy values are no surprise, since also for the test set, the network is fed the correct label and therefore could in principle always return a ``prototypical'' class sample, ignoring all other information. When varying the efficiency loss weight $\alpha_{EF}$, accuracy decreases, but the decrease was only statistically significant for $\alpha_{EF}\geq 8$ (p-value $<$ .001). Adding a discriminator also negatively impacts accuracy with $\alpha_{D}\geq 0.64$ showing statistically significant worse results (p-value $<$ .01).

The reconstruction loss $L_{RE}$ is most tightly correlated with the visual quality of the outcomes. In particular, large AE loss is likely to imply poor visual outcomes, despite the fact that other metrics such as accuracy are indicating good results. This can be observed in Table \ref{tab:res} for $\alpha_{CL}=0.24$. Generally, the reconstruction loss worsens when optimizing for accuracy $\alpha_{CL}>0$ or adding a discriminator $\alpha_{D}\geq 0$. Differences to the baseline are significant (p-value $<$ .01). For adding an efficiency loss differences are only significant for values $\alpha_{EF}\geq 8$ (p-value $<$ .01).

The efficiency loss decreases when adding other losses. For the discriminator differences are not significant compared to the baseline, while for all other losses they are for any value $\alpha_{EF}$  and $\alpha_{CL}\geq 0.1$ (p-value $<$ .01).

\section{Related Work}\label{sec:related work} 
There are numerous types of AE. Related to our applications are denoising AE that are typically used through intentional noise injection with the goal of weight regularization. In contrast, we assume that noise is part of the input data and its removal is thus not motivated by regularization. The idea to combine AEs and GANs for image generation has been explored previously, eg. \cite{bao17} uses a conditional variational AE and applies it for image inpainting and attribute morphing. In this work, we consider a novel application of this architecture type. Our work is a form of image-to-image translation\cite{iso17}. Typically, input and outputs are fairly different, eg. the input could be a colored segmentation of an image not showing any details and the output could be a photo like image with many details. In contrast, in our scenario in- and outputs are fairly similar. For image in-painting or completion \cite{iiz17,yu18} a network learns to fill in blank spaces of an image. In contrast, we might both in-paint and erase. Image manipulation based on user edits has been studied in \cite{zhu2016}. They learn the natural image manifold using a generative adversarial network and express manipulations as constraint optimization problem. They apply both spatial and channel, ie. color, flow regularization. Their primary goal is to obtain realistically looking images after manipulations. Thus, their problem and approach is fairly different. Furthermore, in contrast to the mentioned prior works \cite{iso17,iiz17,yu18,zhu2016,bao17} our work can be classified as unsupervised learning. That is, we do not know the final outputs, i.e. the images that should be proposed to the human. Prior work trains by comparing their outcome to a target. In our case, we do not have pairs of human input (images) and improved input (images) in our training data. 

The field of human-AI interaction is fairly broad. The effect of various user and system characteristics has been extensively studied \cite{rze18}. There has been little work on how to improve communication and prevent misunderstandings. \cite{nicu15} discusses high level, non-technical strategies to deal with errors in communication using speech that originate either from humans or from machines. \cite{bis16} lists some errors that occur when interacting with a robot using natural language, such as grammatical, geometrical misunderstandings as well as ambiguities. \cite{bre05} highlighted the impact of nonverbal communication on efficiency and robustness in communication. It is shown that nonverbal communication can reduce errors. 
\begin{figure}
\vspace{-6pt}
    \setlength\abovecaptionskip{-0.15\baselineskip}
	\centering \includegraphics[width=0.6\textwidth]{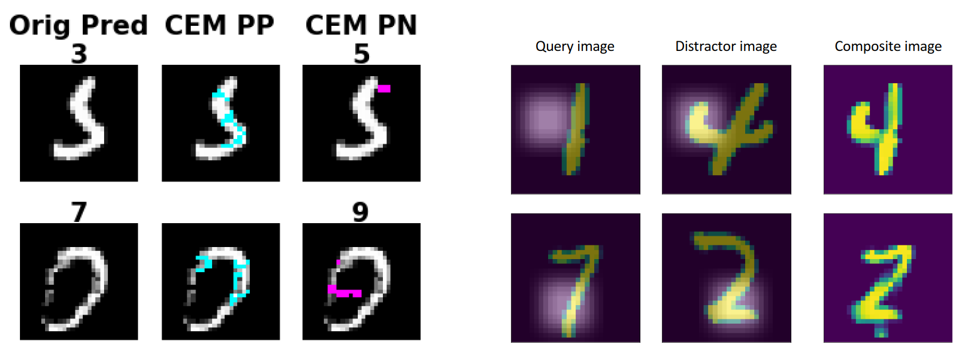}
	\caption{Left digits are taken from \cite{dhu18}. Right digits stem from \cite{goy19}.} \label{fig:otPic}
	\vspace{-6pt}
\end{figure}
Our work also relates to the field of personalized explanations \cite{sch19}. It aims to explain to a user how she might improve interaction with an AI. Explainability in the context of machine learning is generally more focused on interpreting decisions and models (see \cite{ada18,sch19} for recent surveys). Counterfactual explanations also seek to identify some form of modification of the input. \cite{dhu18} explains by answering ``How to modify an input to get classification Y?'' and ``What is minimally needed?''.  The former focuses on mis-classified examples with the goal of changing them with minimal effort to the correct class. For the latter all objectives except efficiency are ignored and there is only the constraint of maintaining classification confidence above a threshold. Thus, \cite{dhu18} discusses special cases of our work. Technically, \cite{dhu18} generates a perturbation added to the sample such that the perturbation is minimal given a threshold confidence of the prediction (either as the correct class or as an alternative class) has been achieved. They use an ordinary AE as an optional element on the perturbation, which does only slightly alter results. In contrast, we use a CCAE on the inputs, which is essential. We optimize for multiple linear weighted objectives without thresholds.  \cite{goy19} aims at explaining counterfactuals, ie. showing how to change a class to another by combining images of both classes. That is, given a query image and a distractor image they generate a composite image that essentially uses parts of each input. For instance, in the right part of Figure \ref{fig:otPic} the ``7'' in the second row serves as query image, the ``2'' in the middle as distractor and the right most column shows the outcome. The implementation relies on a gating mechanism to select image parts. Differences are also noticeable in the outcomes as shown in Figure \ref{fig:otPic}. The highlighted differences appear noisy in \cite{dhu18} and are not necessarily intuitive, eg. for column CEM-PP for digit ``3'' a stroke on top is missing, but \cite{dhu18} finds a miniature ``3'' within the given digit. The generated images in \cite{goy19} appear more natural, but do have artifacts, eg. the ``2'' being a composition of a ``7'' and a ``2'' has a ``dot'' in the bottom originating from the ``7''. In conclusion, while counterfactual explanations \cite{dhu18,goy19} are related to our work, the objectives differ, eg. we include efficiency, as well as methodology and outcomes. While we also make recommendations to a user, there are only weak ties to recommender systems. Even for interpretable recommendation systems \cite{fus19} users typically primarily seek to understand decisions but do not commonly aim to alter their behavior to obtain better recommendations. 

\section{Discussion and Conclusions} \label{sec:conc}
Input from human to AI is likely to gain further in importance. This paper investigated improving information flow from human to AI by proposing adjustments to human generated examples based on optimizing multiple objectives. Our evaluation highlights that such an automatic approach is indeed feasible for handwriting. While we believe that our approach is suitable for other domains such as speech recognition, details of the network architecture, definition of loss terms and the loss weights likely need to be adjusted. Furthermore, our work focused on generating altered input samples fulfilling specific metrics, but it leaves many questions unanswered when applying it. For instance, it did not investigate how these samples are best shown or explained to users, eg. by highlighting differences or, maybe, even in textual form. These points and more advanced multi-objective optimization, ie. exploring the set of (Pareto) optimal solutions rather than manually adjusting parameters $\alpha$, are subject to future work. Furthermore, one might include more objectives, eg. generating proposals that require little energy to process by the AI\cite{sch19Green} or taking into account behavioral norms expected by people as common for social robots\cite{zhao06,bar04}. We hope that in the future human-to-AI coaches will help non-experts to better interact with AI systems.

\bibliography{references}
\bibliographystyle{splncs04.bst}

\end{document}